\documentclass[letterpaper, 10 pt, conference]{ieeeconf}  

\IEEEoverridecommandlockouts                              

\usepackage{fancyhdr,graphicx,amsmath,amssymb}
\usepackage{bm}

\title{\LARGE \bf

Investigating the Usability of Collaborative Robot control through Hands-Free Operation using Eye gaze and Augmented Reality

}

\author{Joosun Lee$^{1}$, Taeyhang Lim$^{2}$ and Wansoo Kim$^{3,*}$
        \thanks{$^{1}$J. Lee is with Department of Mechatronics Engineering, Hanyang University, 55 Hanyangdaehak-ro, Sangnok-gu, Ansan, Gyeonggi-do, 15588, Republic of Korea ({\tt\small e-mail:km01049@hanyang.ac.kr}).}
        \thanks{$^{2}$T. Lim is with Department of Interdisciplinary Robot Engineering Systems, Hanyang University, Ansan, Gyeonggi-do, Republic of Korea.}
        \thanks{$^{3}$W. Kim is with Robotics Department, Hanyang University ERICA, Ansan, Gyeonggi-do, Republic of Korea ({\tt\small e-mail:wansookim@hanyang.ac.kr}).}
        \thanks{$^{*}$Corresponding Author }
}

\begin{document}
\maketitle
\thispagestyle{empty}
\pagestyle{empty}

\begin{abstract}

This paper proposes a novel operation for controlling a mobile robot using a head-mounted device. Conventionally, robots are operated using computers or a joystick, which creates limitations in usability and flexibility because control equipment has to be carried by hand. This lack of flexibility may prevent workers from multitasking or carrying objects while operating the robot. To address this limitation, we propose a hands-free method to operate the mobile robot with a human gaze in an Augmented Reality (AR) environment. The proposed work is demonstrated using the HoloLens 2 to control the mobile robot, Robotnik Summit-XL, through the eye-gaze in AR. Stable speed control and navigation of the mobile robot were achieved through admittance control which was calculated using the gaze position. The experiment was conducted to compare the usability between the joystick and the proposed operation, and the results were validated through surveys (i.e., SUS, SEQ). The survey results from the participants after the experiments showed that the wearer of the HoloLens accurately operated the mobile robot in a collaborative manner. The results for both the joystick and the HoloLens were marked as easy to use with above-average usability. This suggests that the HoloLens can be used as a replacement for the joystick to allow hands-free robot operation and has the potential to increase the efficiency of human-robot collaboration in situations when hands-free controls are needed.


\end{abstract}

\section{INTRODUCTION}



Human-robot collaboration involves the sharing of tasks between humans and robots, and is commonly used in industrial environments where humans operate robots to carry heavy loads or perform repetitive tasks. Depending on the roles of the human and the robot, they devise ways to interact with each other \cite{intro_1}. This collaborative work allows for increased efficiency, as robots' advantages in accuracy and strength can be combined with humans' flexibility and adaptability \cite{wang2020overview}.

One of the main goals in physical human-robot collaboration is for the robot to understand and interpret human intentions at various stages of collaborative tasks. This allows the robot to generate appropriate responses in collaborative settings. By reflecting human intentions and states, the robot's actions are determined to ensure the smooth progress of the task \cite{intro_2}.

\begin{figure}[t!]
    \centering
    \includegraphics[trim=0.0cm 1.0cm 0.0cm 1.5cm,clip,width=\linewidth]
    {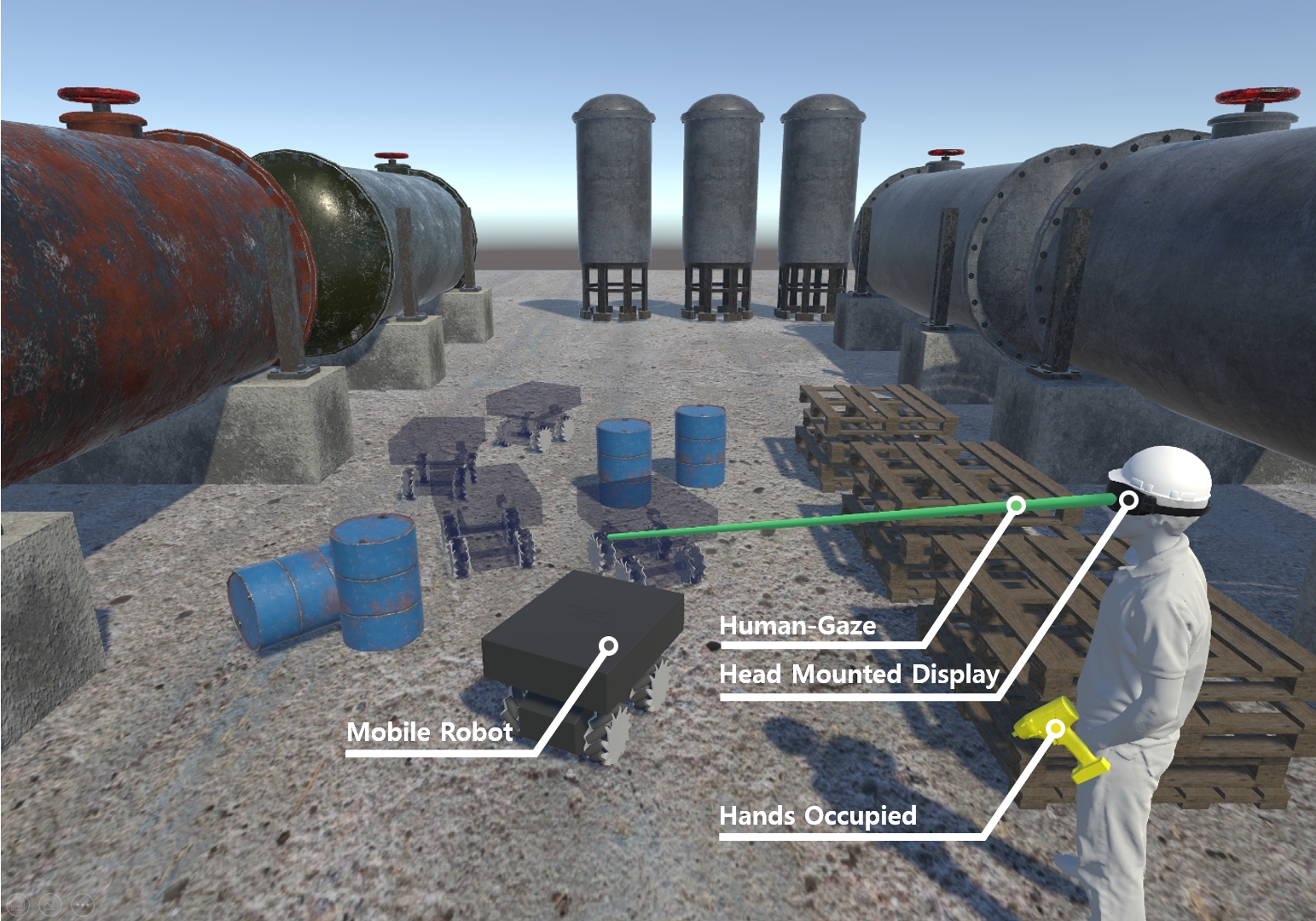}
    \caption{An illustration of the use of Hololens for human-robot interaction in a worksite. The mobile robot moves based on the human's visual perception using gaze only. This allows for the natural and intuitive operation of the robot in a hands-free manner.}
    \label{fig:concept}
    \vspace{-0.5cm}
\end{figure}

There are various ways to interpret human intention. Commonly, human intentions are delivered to the robot through physical controls such as a computer or joystick \cite{intro_8}\cite{intro_9}. However, previous studies have shown that nonverbal behaviors are often quicker than verbal behaviors, as people tend to look before acting \cite{land}. One common way to interpret nonverbal behavior is through the human gaze. Gaze information is useful as it does not place any additional burden on humans compared to methods such as verbal or computer commands. For such reasons, it is widely used not only in medical and engineering fields, but also in psychology \cite{intro_3}.


Augmented reality (AR) has been adopted in many industries to improve efficiency, safety, and quality in various processes \cite{gkournelos2018application}\cite{evans2017evaluating}\cite{9473546}. One of the advantages of using AR is its ability to enhance safety. With the help of HoloLens devices, workers can monitor the safety of their tasks and surroundings to prevent accidents \cite{wilson2005design}. Therefore, AR is a suitable solution for industries where safety is a priority, such as manufacturing and construction. Additionally, AR can improve the way assembly line tasks are performed. By providing 3D models of parts through an AR interface, workers can visualize the assembly process and perform tasks more accurately \cite{lamon2019capability}. This has been particularly useful in shared work environments, such as car assembly lines, where workers can collaborate more effectively and integrate their tasks \cite{dalle2021augmented}. Hence, AR enables physical interaction between humans and robots in the workspace in a safe and efficient way.

In this paper, we propose a novel for hands-free human-robot interaction of a mobile robot using a head-mounted device (HMD). From the prior studies mentioned above, it is evident that the human gaze can be used to interpret human intention and AR can be used to safely operate the robot. Thus, a collaborative approach for smooth communication between humans and robots using gaze and AR is proposed. Through the use of the Microsoft HoloLens 2, real-time gaze data is delivered to the virtual robot model to determine the gazed location and direction. The real robot is navigated simultaneously with the virtual robot and is controlled through the admittance control algorithm. Through admittance control, the forces are exploited to the robot after interpreting human intention. Furthermore, we perform the experiments to evaluate the usability and efficiency of the proposed method, comparing it to traditional methods of robot operation such as using a joystick.

The rest of the paper is structured as follows. In section II, the related works are introduced. In section III, the proposed hands-free operation for mobile robot is explained. In 
section VI, the experiments are described and the results obtained from the robot experiments are presented. Finally, conclusions and future work are discussed in section V.

\section{RELATED WORK}

\begin{figure}
    \centering
    \includegraphics[trim=0cm 0cm 0cm 0cm,clip,width=1\linewidth]{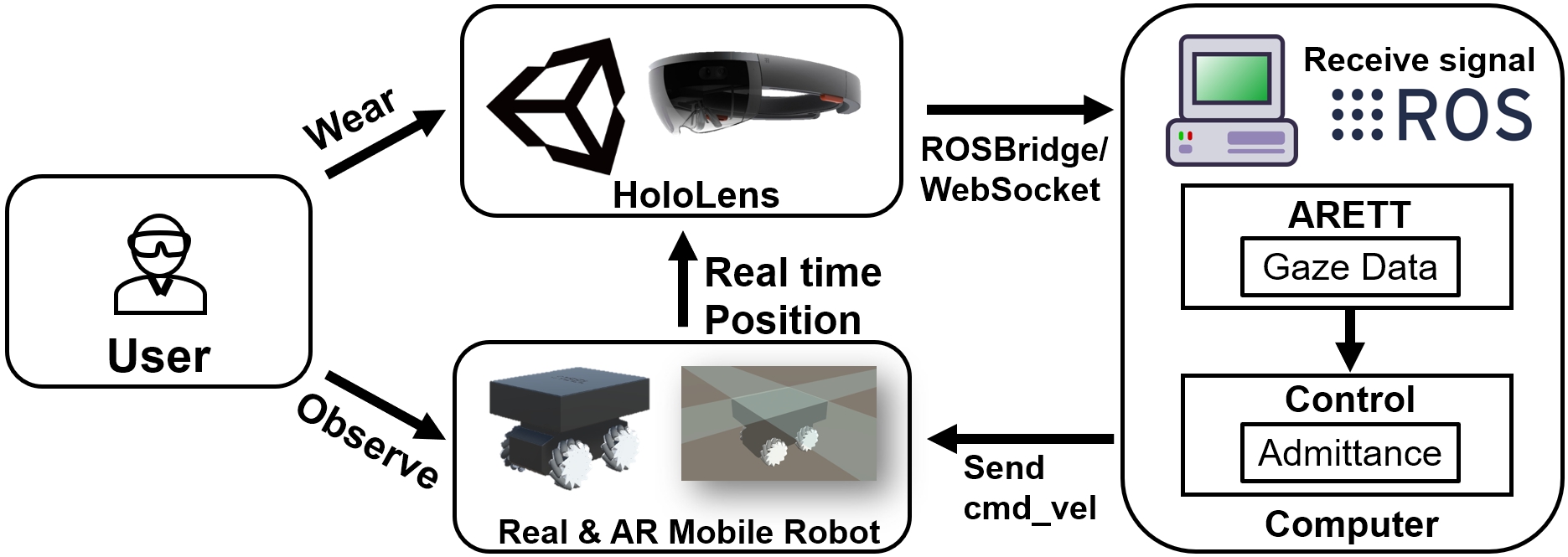}
    \caption{Framework of the components in the proposed method.}
    \label{fig:frame}
    \vspace{-0.5cm}
\end{figure}

Several studies have demonstrated that promoting human-robot interaction in a shared workspace can enhance performance and efficiency \cite{kosuge2004human}\cite{scholtz2003theory}\cite{lasota2017survey}. 
The joystick is a common control method for robots as it provides direct input to manipulate the robot's actions. However, its limitation of requiring hands-on operation may hinder users from carrying other objects or prevent flexible human-robot collaboration such as role assignment, co-carrying tasks, etc.  
To allow hands-free operation, the HMD is an alternative way to control the robot.

HMDs include augmented reality (AR) devices that display computer-generated images in a real environment \cite{AR} and virtual reality (VR) devices that immerse the user in a simulated environment \cite{mine1995virtual}.
For robot controls in the real environment, AR is a preferable solution because the user can easily adapt and observe the external environment as the real and virtual objects are displayed simultaneously.
Additionally, an important part of this paper is to observe human intention through the eye gaze. There are currently various HMDs available for gaze tracking \cite{cognolato2018head}. To fulfill the requirement of both AR display and gaze tracking in a single head-mounted device, Microsoft HoloLens 2 is selected. HoloLens is a smart glasses device developed by Microsoft that supports augmented reality (AR) \cite{holo} and has the advantage of being able to measure human gaze. 

Previously research by \cite{joy} combined a joystick and HoloLens together to operate real robots. The research exploited teleoperation where the user was able to visualize the robots' perspective through the HoloLens along with a 3D virtual robot model. The user operates the virtual robot using a joystick to operate a real robot that is located in another place. However, the joystick restricts the users' ability to hold other materials and so it is not ideal for multitasking. For such reasons, the proposed research aims to use gaze within the HoloLens to operate the robot rather than a joystick. 

Although HoloLens can track human gaze data, the data is not accessible by the developers. 
The augmented reality eye-tracking toolkit (ARETT) provides a solution to collect and store eye-gaze data \cite{ARETT}, but it cannot access real-time data to be used to manipulate the robot. Thus, modification of the toolkit is required to access real-time gaze data. 

\begin{figure}
    \centering
    \includegraphics[trim=1cm 0.3cm 0cm 0.2cm,clip, width=\linewidth]{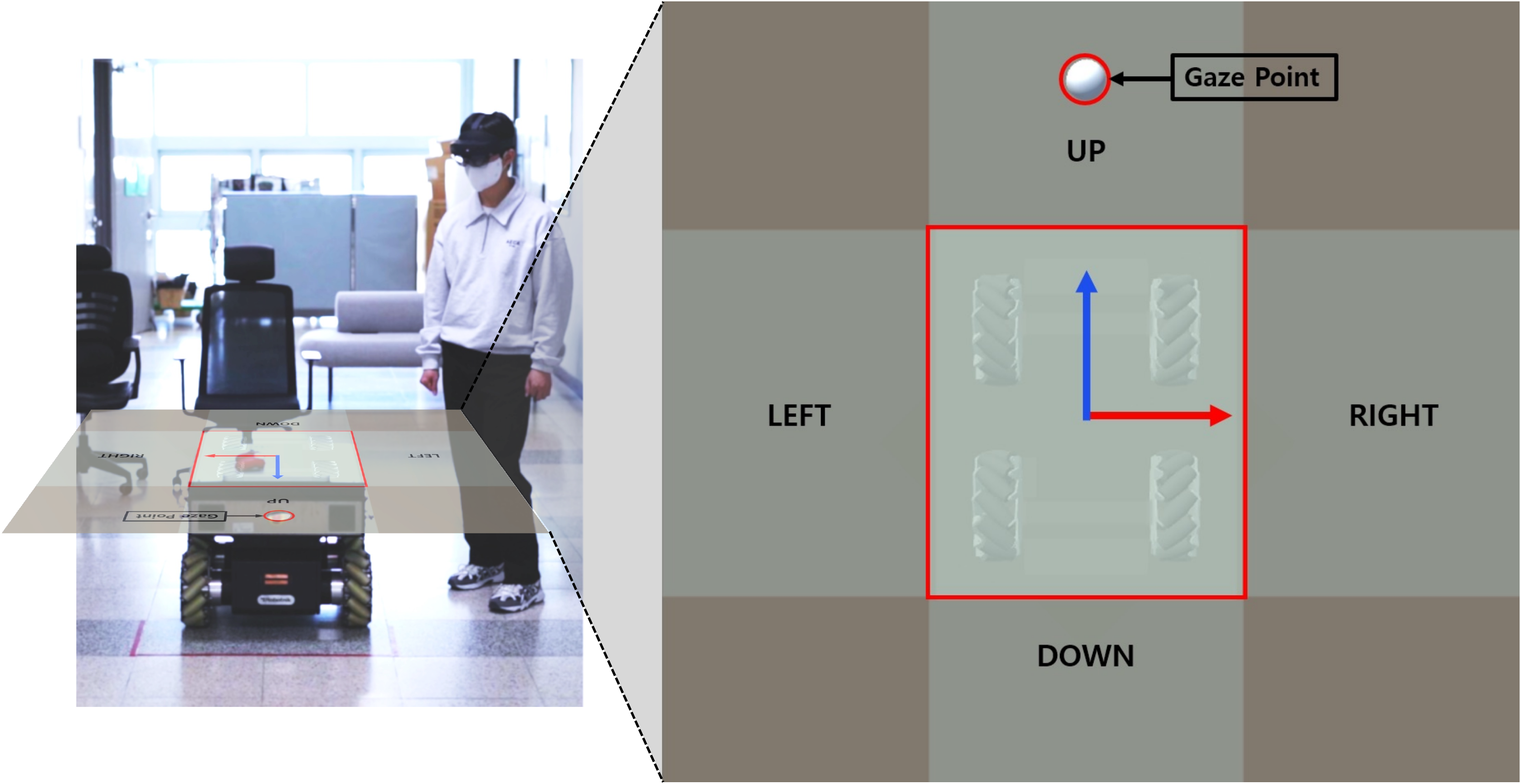}
    \caption{Augmented reality display of the mobile robot (Left). Gaze point is represented as a white sphere. When gaze point lies on the up, down, left or right square, the mobile robot moves accordingly.}
    \label{fig:Gaze}
    \vspace{-0.5cm}
\end{figure}

This study aims to develop a human-robot interaction for controlling a mobile robot through hands-free operation using real-time eye-tracking data and a virtual robot model based on HoloLens 2. A system that leverages the Mixed Reality app on HoloLens 2 is developed to provide a real-time interface for controlling a mobile robot. The system captures eye-tracking data in real-time and then uses it to generate navigation commands for a mobile robot through the admittance control algorithm.

The research highlights the potential of HoloLens and mixed reality apps to enhance the control and interaction capabilities of mobile robots. Real-time eye-tracking data and virtual robot model enable more natural and intuitive control of robot, which can be especially useful in settings where traditional input devices (i.e., joysticks or keyboards) are not practical or effective.

\section{Method}

The proposed method allows for the safe control of a mobile robot using eye tracking. It combines both the real and AR environments to implement collaborative human-robot interaction. The overall framework of the proposed method is shown in Fig. \ref{fig:frame}. The user wears a HoloLens to match the positions of both the real and virtual robots and transfers their visual data to the computer through rosbrige. The computer then combines the eye-gaze data with the robot position information to enable stable control of the mobile robot. The user can control the mobile robot by looking at both the real and virtual robots simultaneously.

\subsection{Augmented Reality and Eye Gaze Tracking System}

The Microsoft HoloLens 2 Development Edition is used to capture human gaze data and display augmented reality (AR) \cite{holo}. The Unity 3D game development engine is used to create the scene for robot operation, with the robot model and control display represented as AR objects (see Fig. \ref{fig:Gaze}). The mixed reality toolkit provided by Microsoft allows for the use of various sensors built into the HoloLens, including head, eye, hand, and voice tracking. In this work, only the eye tracking function is employed to localize the gaze location for hands-free control.

To synchronize the virtual and real robots, the starting position of the HoloLens (0.5m behind the real robot) is used. Both mobile robots are manipulated using the same speed and directional changes.

Although HoloLens 2 enables gaze tracking, the device does not provide access to gaze data. To address this, the ARETT toolkit \cite{ARETT} is integrated into the system to obtain gaze data from the HoloLens 2. However, as ARETT stores gaze data rather than providing it in real-time, the proposed study modifies the toolkit to obtain real-time gaze data. The gaze data is processed at a rate of 30Hz (i.e., every 33.33ms).




\begin{figure}[t!]
    \centering
    \includegraphics[trim=0.0cm 0.2cm 0.0cm 0.2cm,clip,width=5cm, height=4cm]
    {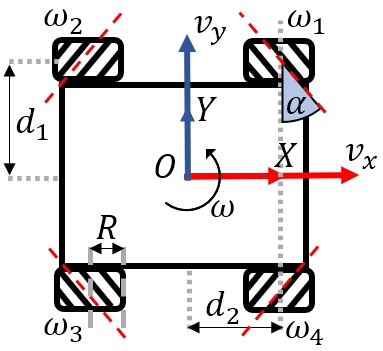}
    \caption{Kinematic modeling of a mecanum-wheeled robot. $d$ represents the distance, $w$ represents the wheel rotation and $\alpha$ represents the angle between the wheel axis and the roller.}
    \label{fig:mobile}
    \vspace{-0.2cm}
\end{figure}




\begin{figure}[t!]
    \centering
    \includegraphics[trim=0.0cm 0.0cm 0.0cm 0.0cm,clip,width=9cm, height=2cm]
    {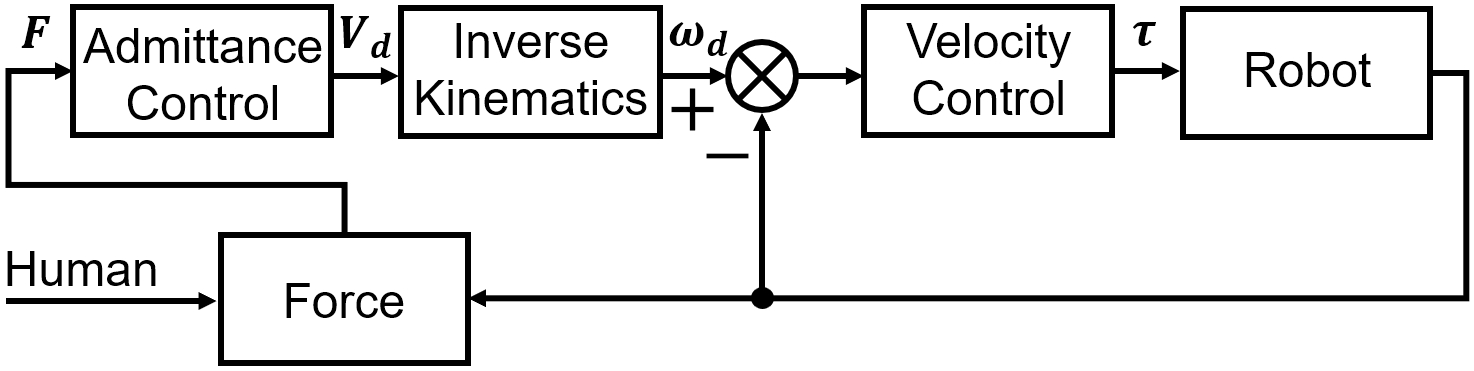}
    \caption{Overview of the systematic structure of the admittance based control. $F$ represents the force, $V$ represents the velocity, $w$ represents the rotation for each wheel and $\tau$ is the time constant.}
    \label{fig:Admittance}
    \vspace{-0.5cm}
\end{figure}

\subsection{Mobile Robot Kinematic}
In this work, we use a commercial mobile robot, namely Summit XL Steel, which is designed for industrial applications. It has a dimension of 750 $\times$ 665 $\times$ 510mm with a payload of up to 100kg. It has four mecanum wheels which are designed to allow robots to move in any direction without changing direction Omni-directional base (see Fig.\ref{fig:mobile}). 
The velocity of the mobile robot $\bm{V} = \begin{bmatrix}\label{eq1} v_{x}& v_{y}& \omega \end{bmatrix} \in \mathrm{R}^3$  is represented as follows:
\begin{equation}
    \bm{V}=-\frac{R}{4}\begin{bmatrix} -\tan\alpha & \tan\alpha & -\tan\alpha & \tan\alpha\\ 1 & 1 & 1 & 1\\ L & -L & -L & L \end{bmatrix} \bm{\omega}_{H}, \ 
    \end{equation} 
where $L = \frac{1}{d_{1}+d_{2}\cot\alpha}$, $d_{1}$ and $d_{2}$ represent the distance from the center of the robot along the X-axis and Y-axis, respectively, $R$ is the radius of the wheel and $\alpha = 45$ is the angle between the wheel axis and the roller. 
The vector $\bm{\omega}_{H} = \begin{bmatrix}\omega_{1}& \omega_{2}& \omega_{3}& \omega_{4}\end{bmatrix}^{T}$ represents the speed of each wheel that are individually driven by four motors, it can be obtained as
\begin{equation}
    \bm{\omega}_{H}=\begin{bmatrix} -\cot\alpha & 1 & (d_{1}+d_{2}\cot\alpha)\\ \cot\alpha & 1 & -(d_{1}+d_{2}\cot\alpha)\\ -\cot\alpha & 1 & -(d_{1}+d_{2}\cot\alpha)\\ \cot\alpha & 1 & (d_{1}+d_{2}\cot\alpha) \end{bmatrix}\bm{V}.\label{eq3}
\end{equation}
The kinematics equations for both forward and inverse movements of the mecanum wheeled robot are used in the robot control process.




\subsection{Admittance Behaviour for Hands-free Operation}

Admittance control is a method of using the robot's position and force measurements for control \cite{Admittance}. Through the forces and movement measured in the environment, the robot adjusts its movement in response to the environment. Admittance control can increase the stability and accuracy of robot operation depending on the environment. From this, the robot can safely adapt to different environments and enables safer and smoother collaboration between robots and humans. 


This study utilizes an admittance model as the transfer function for a human-robot system.
To take into account the user's intended movement, we define a dead-zone within the mobile platform in AR, which is treated as a no-movement area for the mobile robot.

The virtual force $\bm{F}$ serves as the input of admittance control and corresponds to the user's intention. Accordingly, we compute the displacement of the eye gaze position $\bm{G}_{P}$ w.r.t the virtual robot position $\bm{R}_{P}$. The displacement is set to zero if the gaze position is placed inside of the dead-zone, while it equal to the distance between $\bm{G}_{P}$ and $\bm{R}_{P}$. The result of displacement is then used to compute the virtual force $\bm{F}$ by:

\begin{equation}\label{eq5}
    \bm{F} = K({\bm{G}_{P}} - {\bm{R}_{P}}),
\end{equation}
\noindent where ${K}$ is the virtual stiffness parameter, which is experimentally chosen based on the safety.

This virtual force is now used to obtain the desired velocity ${\bm{{V}}_{d}(t)}$:
\begin{equation}\label{eq4}
    {\bm{{V}}_{d}(t)}=\frac{\bm{{F}}}{{D}}(1-e^{-t/\tau}) ,
\end{equation}

\noindent where the time constant of the system is denoted as $\tau$ and is defined by the ratio of the virtual mass coefficient, ${M}$, to the virtual damping coefficient, ${D}$ (i.e., $\tau$ = $M$/$D$). The steady-state velocity of the system, $\bm{V}$, is given by dividing the force $\bm{F}$ by ${D}$. This means that the velocities of the human-machine system are determined by the forces exerted by the user onto the robot. For example, when the user's steady forward walking velocity is $\bm{V}$, the required pushing force $\bm{F}$ or the burden that the user feels reacted from the robot, should be adjusted accordingly by altering the virtual damping coefficient $D$. Additionally, by changing the virtual mass coefficient $M$ (thus altering $\tau$), different dynamic responses of the human-machine system can be obtained.

\begin{figure}[t!]
    \centering
    \includegraphics[trim=0.0cm 0.0cm 0.0cm 0.0cm,clip,width=0.95\linewidth]
    {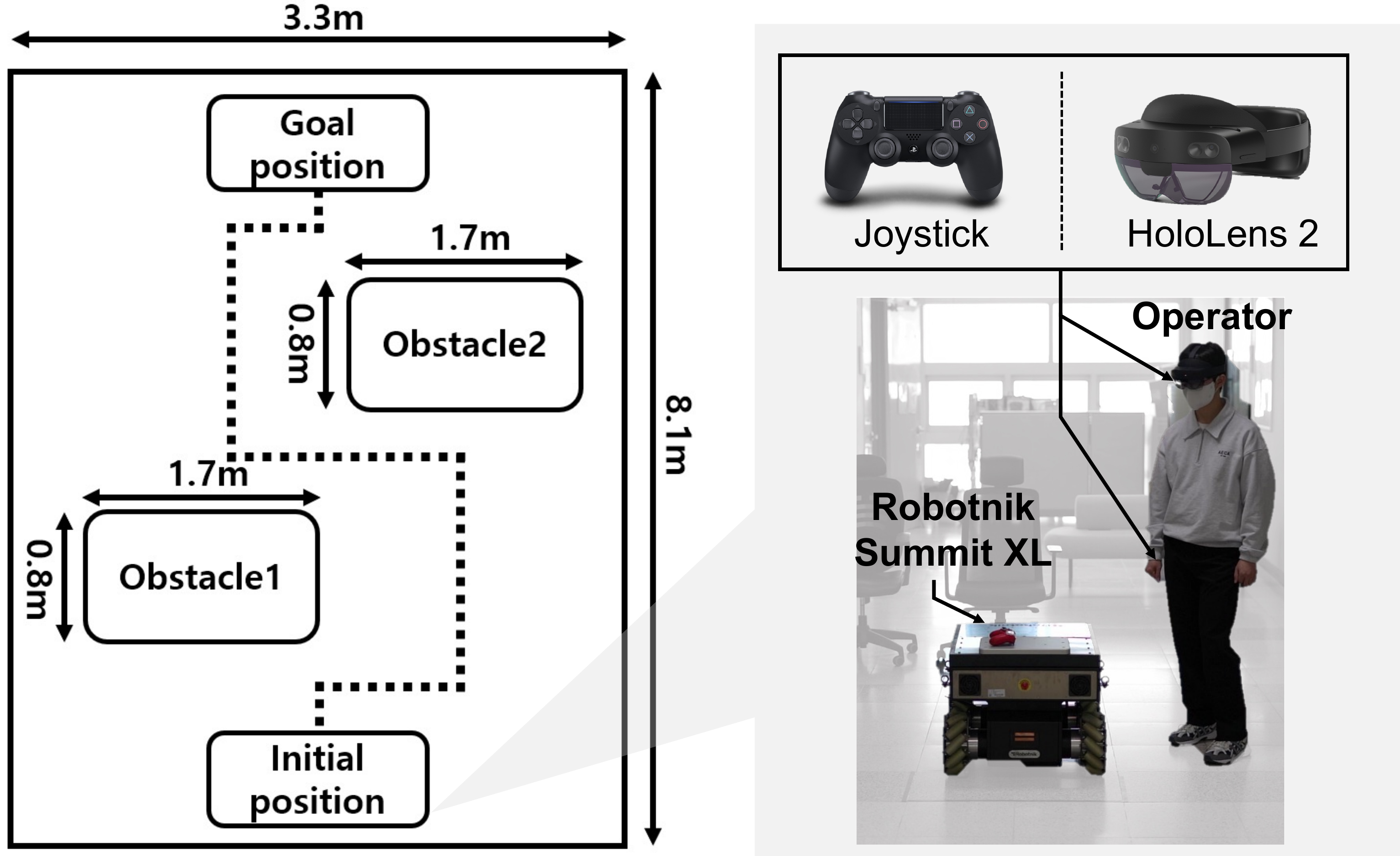}
    \caption{The experimental setup for the robot control experiments. The setup was identical for the comparison experiment performed by the joystick and the HoloLens.}
    \label{fig:env}
    \vspace{-0.5cm}
\end{figure}

The admittance-based control method is illustrated in Fig. \ref{fig:Admittance}. Consequently, the target driving velocities $\bm{\omega}_{H|d}$ for each wheel can be obtained by Eq.\eqref{eq3}  with input as ${\bm{{V}}_{d}(t)}$, and $\bm{\omega}_{H|d}$ for each wheel is used for velocity control.

\subsection{Robot Operating System (ROS)}

ROS (Robot Operating System) is an open-source framework for robot software development. It allows multiple processes to communicate with each other through a messaging system. ROS is used in a wide range of robot applications, including autonomous vehicles, industrial automation, and service robots.


The streamed data from the HoloLens gets communicated to the Robot Operating System (ROS). Additional to the gaze data, the real robot and Virtual robot position and rotation get delivered through ROS communication protocol. The data get used to operate the real robot to act accordingly to the gaze direction. The communication is made possible through a wireless local area network (WLAN) and the use of rosbridge.

\section{EXPERIMENTS \& RESULTS}

\begin{figure}[t!]
    \centering
    \includegraphics[trim=3.0cm 0.5cm -0.8cm 0.0cm,clip,width=10cm, height=4.5cm]
    {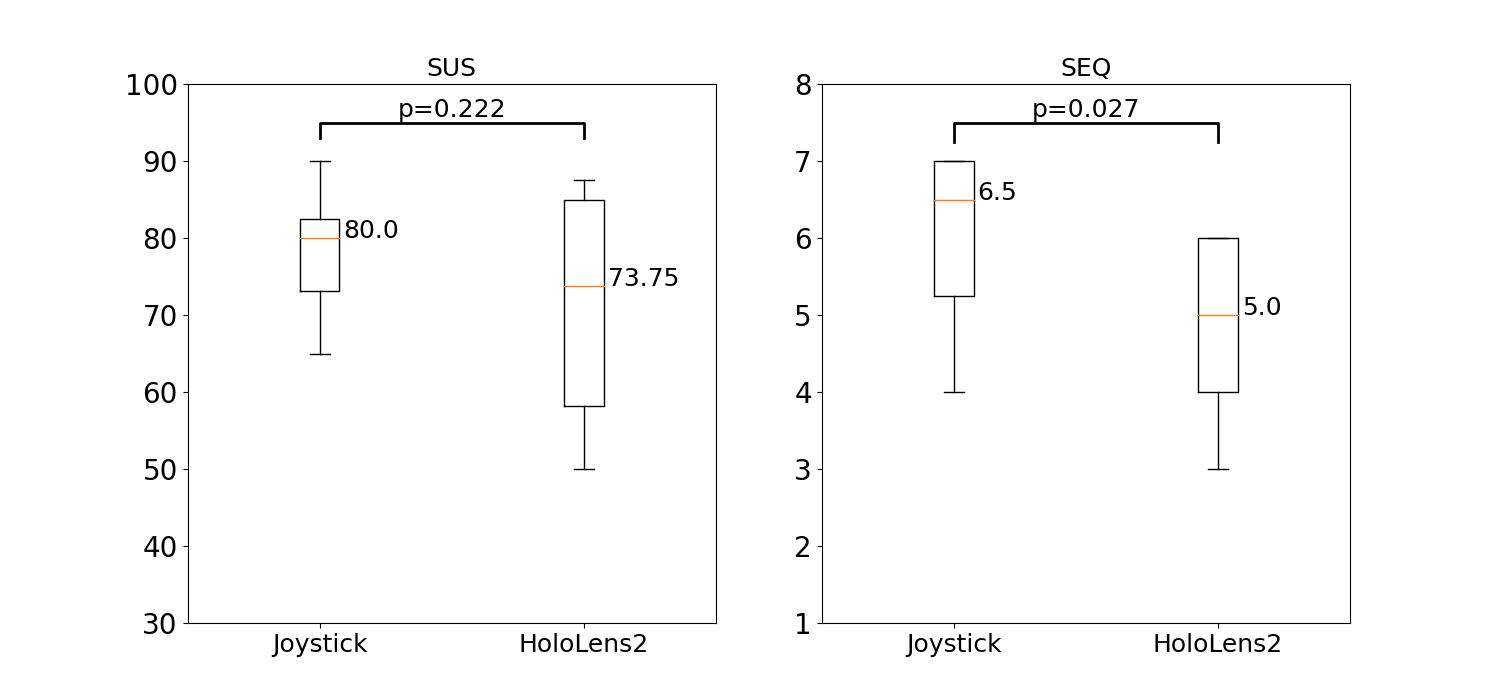}
    \caption{Questionnaire score summary: (left) system usability scale (SUS) and (right) single ease question (SEQ).}
    \label{fig:SUSSEQ}
    \vspace{-0.2cm}
\end{figure}

In this section, we provide the experimental validation of the proposed framework's usability from the hands-free operation method, where a mobile robot was operated through intuitive control intention using the user's eye gaze, not hands. 

\subsection{Experimental Setup}
During the experiment, the HoloLens 2 was used to collect human gaze data in AR. The data from the HoloLens was transmitted in real-time to the PC using the ROS web and Rosbridge for controlling the real mobile robot. 
The mobile robot (Robotnik Summit XL) was used to execute the admittance behavior by means of the user's eye gaze.
The virtual mobile robot was used as a reference to control the real mobile robot. To synchronize the virtual robot model with the real robot, the HoloLens was started 0.5m behind the real robot. Thus, the virtual robot model was positioned in the same way as the actual robot. The force of the gaze was calculated only in the direction of movement. Thus when moving up and down, only the force on the x-axis was considered. And when moving left and right, only the force on the y-axis was considered. The working environment is shown in Fig. \ref{fig:env}. The pathway to the destination had several obstacles so that the participant cannot see the target point from the starting position. Thus, the participant had to follow the mobile robot to reach the target. 

There were a total of ten participants in the experiment, with six males and four females. Prior to conducting the experiment, instructions were provided and the participants were instructed on how to control the mobile device using HoloLens 2. Gaze calibration was also performed using the calibration set provided on the HoloLens. All participants experienced the AR system for 2 minutes to become familiar with it.

\begin{figure*}
    \centering
    \includegraphics[trim=0cm 0cm 0cm 0cm,clip,width=1.0\linewidth]
    {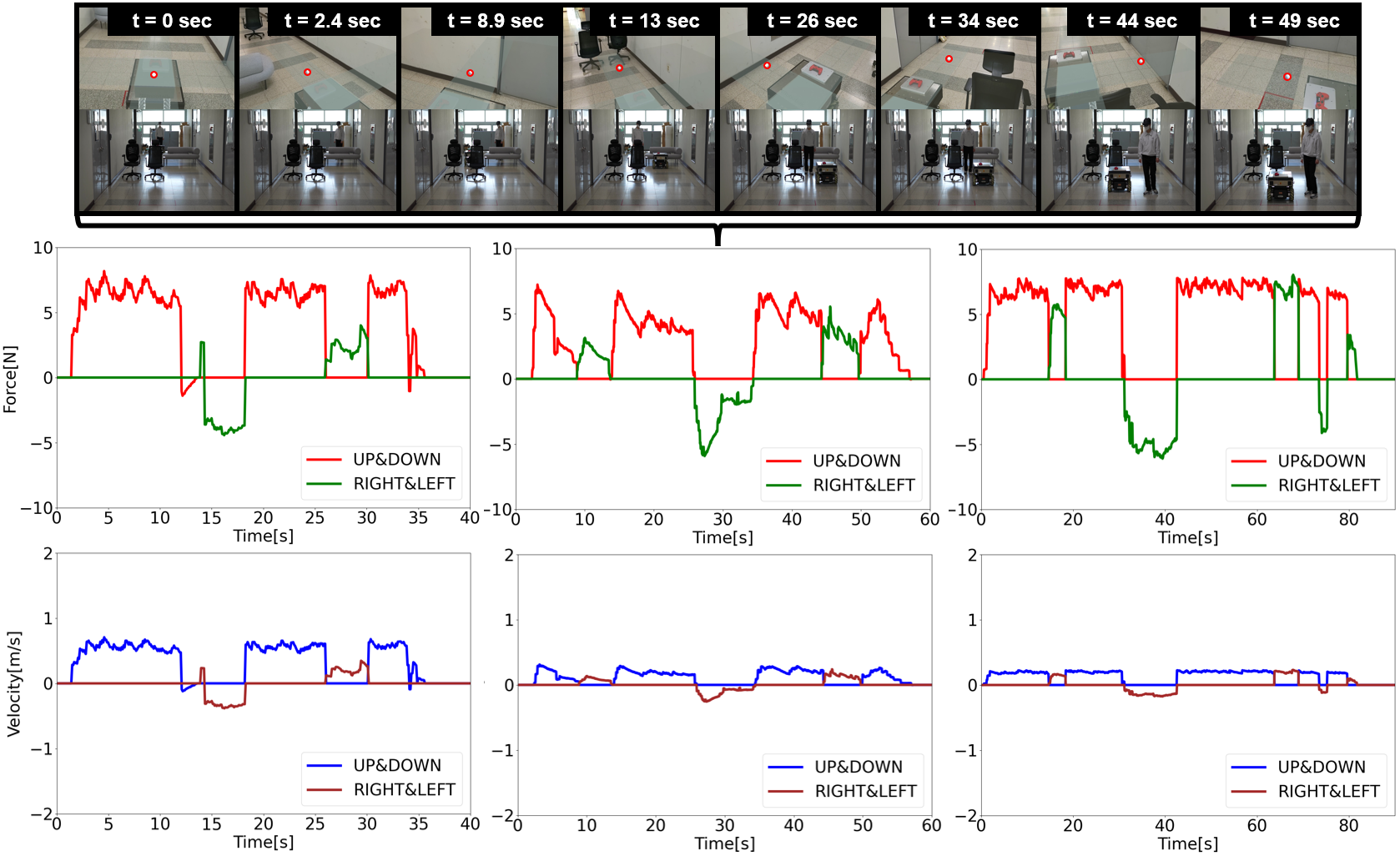}
    \caption{Snapshots of a HoloLens experiment with damping of 20 N$\cdot$s/m (top) where the gaze point is indicated by a red circle. The Force (middle) and the velocity (bottom) are illustrated when tested with damping of 10 (left), 20 (middle) and 30 N$\cdot$s/m (right).}
    \label{fig:sequence}
    \vspace{-0.5cm}
\end{figure*}


In a complementary hands-free operation experiment, we compared two different operation methods (i.e., with joystick and with HoloLens) for controlling the mobile robot. Therefore, a total of four experiments were conducted per subject. 
The first experiment did not require the use of HoloLens. The participants were asked to reach the destination using a joystick. The maximum robot speed using the joystick was 0.5 m/s. After, the following three experiments were conducted through the HoloLens without the joystick, where the robot had different damping values of 10, 20 and 30 N$\cdot$s/m, respectively. 
The participant had 1 minute of rest time after each experiment. 
To investigate of the proposed method, we first provided the results of robot control performance, then the usability survey was conducted after each experiment. A statistical method was employed to compare the significant differences between them (level of statistical significance used was 0.05).
The objective was to demonstrate a high usability across all users when hands-free operation was employed.

\subsection{Robot Control Results}

The result of hands-free operation experiments with different robot damping values are illustrated in Fig. \ref{fig:sequence}. Top row in Fig. \ref{fig:sequence} shows Eight sequences of hands-free experiment under the damping condition of 20 N$\cdot$s/m. The middle and bottom row in Fig. \ref{fig:sequence} show the virtual force and the desired velocity with three different damping as 10, 20, and 30 N$\cdot$s/m, respectively. 
The force and velocity values increased when the user looked in the direction of intention while the admittance control ensured robot stability. The lowest damping value of 10 N$\cdot$s/m resulted in the fastest reaching of the destination due to the admittance controller.

From the four experiments, the time to reach the destination was measured and were averaged across the users. Through the joystick, the time to reach the target was 55s, whereas using the HoloLens, it took average of 45.76s, 58.2s, and 74s  with damping of 10, 20, and 30 N$\cdot$s/m, respectively. This shows that the time difference between the joystick and HoloLens depends on the damping values as using the HoloLens could result in slower and faster target reach compared to the joystick.


Additionally, a simple post-task question (``Which damping value for the mobile robot provided the most comfortable experience?'') measuring user's preference. The result indicates that the participants preferred a damping value of 20 N$\cdot$s/m the most. Also, the users left additional opinions on their experience using the HoloLens in replacement of the joystick. The comments were positive as one of the participants noted ``I experienced the convenience of being able to control a robot by simply using my eyes without any additional actions.''

\subsection{Usability Test Results}
Two indices were used to evaluate the usability of the proposed method in terms of cognitive aspects: System Usability Scale (SUS) and Single Ease Question (SEQ). Pairwise comparisons were conducted by using post-hoc paired t-tests.
SUS is a commonly used survey to evaluate the usability of a system \cite{brooke1996sus}. It consists of 10 questions with 5-point responses ranging from ``strongly agree'' to ``strongly disagree.'' An above-average SUS score for usability is 73.2, while a SUS score between 68 and 73.2 is considered indicative of a usable product. SUS was employed to evaluate the user whether the HoloLens is as intuitive as the joystick. On the other hand, SEQ is a useful tool for evaluating usability in various fields, such as comparing and evaluating measurement tools based on user experience \cite{SEQ}. The SEQ was acquired from every participant to indicate the level of ease or difficulty experienced by the user while performing with different devices.

Fig. \ref{fig:SUSSEQ} presents the overall experimental results of the usability test. In the left plot on Fig. \ref{fig:SUSSEQ} depicts the SUS result. The average SUS score of both the joystick and the HoloLens was above-average in its usability. Where the average score for the joystick was 80.0 (SD=12.5) and the HoloLens was 73.75 (SD=18.75). Statistical results show no significant differences ($p$-value = 0.22).
This is mainly due to the unfamiliarity with a new device. Nevertheless, the average SUS score was similar between the two devices, it indicates that the HoloLens is equally user-friendly when compared to the joystick.



In the right plot on Fig. \ref{fig:SUSSEQ} presents the results of SEQ scores.
The conventional joystick method resulted in 6.5 points (SD=1.5) and the HoloLens method resulted in 5 points (SD=1.5). The two methods were shown to be statistically significant ($p$-value $<$ 0.05).
This shows that although the participants may have shown a preference for the joystick in the experiment, the results show that the use of the HoloLens was still easy to use, as reflected by the score of 5 points. 
This demonstrates the overall usability of the proposed method for eye gaze based mobile robot control under the condition of admittance control manner.


\section{CONCLUSIONS}
This paper presented a way to intuitively convey a human's intended position to a robot using only the user's eye gaze. When the user and mobile robot were existed in the same environment, AR interface was employed to improve the intuitiveness to control the robot through eye gaze. 
An admittance controller was implemented to transfer human-intended force (based on eye gaze) to mobile robot trajectories. 
We experimentally validated performance and usability of the presented framework. The use of the HoloLens resulted in above average in its usability. Thus, human-robot collaboration through a hands-free method using the gaze is a promising solution to operate robots. Consequently, this paper provides valuable insights into the potential of using eye gaze as an intuitive method for human-robot collaboration. Future work will focus on addressing synchronization issues caused by mobile slippage to further enhance accuracy and effectiveness using external camera. 



\addtolength{\textheight}{0cm}   

\section*{ACKNOWLEDGMENT}

This paper was supported by Korea Institute for Advancement of Technology (KIAT) grant funded by the Korean Government (MOTIE). (P0012744, HRD program for industrial innovation). and the National Research Foundation of Korea(NRF) grant
funded by the Korea government(MSIT) (No. 2022R1C1C1008306).

\bibliographystyle{ieeetr}
\bibliography{Ref.bib}

\begin{thebibliography}{10}

\bibitem{intro_1}
A.~Ajoudani, A.~M. Zanchettin, S.~Ivaldi, A.~Albu-Sch{\"a}ffer, K.~Kosuge, and
  O.~Khatib, ``Progress and prospects of the human--robot collaboration,'' {\em
  Autonomous Robots}, vol.~42, pp.~957--975, 2018.

\bibitem{wang2020overview}
L.~Wang, S.~Liu, H.~Liu, and X.~V. Wang, ``Overview of human-robot
  collaboration in manufacturing,'' in {\em Proceedings of 5th International
  Conference on the Industry 4.0 Model for Advanced Manufacturing: AMP 2020},
  pp.~15--58, Springer, 2020.

\bibitem{intro_2}
K.~Sakita, K.~Ogawara, S.~Murakami, K.~Kawamura, and K.~Ikeuchi, ``Flexible
  cooperation between human and robot by interpreting human intention from gaze
  information,'' in {\em 2004 IEEE/RSJ International Conference on Intelligent
  Robots and Systems (IROS)(IEEE Cat. No. 04CH37566)}, vol.~1, pp.~846--851,
  IEEE, 2004.

\bibitem{intro_8}
B.~M. Faria, L.~Ferreira, L.~P. Reis, N.~Lau, M.~Petry, and J.~Couto, ``Manual
  control for driving an intelligent wheelchair: A comparative study of
  joystick mapping methods,'' {\em environment}, vol.~17, p.~18, 2012.

\bibitem{intro_9}
B.~M. Faria, L.~M. Ferreira, L.~P. Reis, N.~Lau, and M.~Petry, ``Intelligent
  wheelchair manual control methods: A usability study by cerebral palsy
  patients,'' in {\em Progress in Artificial Intelligence: 16th Portuguese
  Conference on Artificial Intelligence, EPIA 2013, Angra do Hero{\'\i}smo,
  Azores, Portugal, September 9-12, 2013. Proceedings 16}, pp.~271--282,
  Springer, 2013.

\bibitem{land}
M.~F. Land and S.~Furneaux, ``The knowledge base of the oculomotor system,''
  {\em Philosophical Transactions of the Royal Society of London. Series B:
  Biological Sciences}, vol.~352, no.~1358, pp.~1231--1239, 1997.

\bibitem{intro_3}
D.~Trombetta, G.~S. Rotithor, I.~Salehi, and A.~P. Dani, ``Human intention
  estimation using fusion of pupil and hand motion,'' {\em IFAC-PapersOnLine},
  vol.~53, no.~2, pp.~9535--9540, 2020.

\bibitem{gkournelos2018application}
C.~Gkournelos, P.~Karagiannis, N.~Kousi, G.~Michalos, S.~Koukas, and S.~Makris,
  ``Application of wearable devices for supporting operators in human-robot
  cooperative assembly tasks,'' {\em Procedia CIRP}, vol.~76, pp.~177--182,
  2018.

\bibitem{evans2017evaluating}
G.~Evans, J.~Miller, M.~I. Pena, A.~MacAllister, and E.~Winer, ``Evaluating the
  microsoft hololens through an augmented reality assembly application,'' in
  {\em Degraded environments: sensing, processing, and display 2017},
  vol.~10197, pp.~282--297, SPIE, 2017.

\bibitem{9473546}
M.~Walker, H.~Hedayati, J.~Lee, and D.~Szafir, ``Communicating robot motion
  intent with augmented reality,'' in {\em 2018 13th ACM/IEEE International
  Conference on Human-Robot Interaction (HRI)}, pp.~316--324, 2018.

\bibitem{wilson2005design}
J.~Wilson, D.~Steingart, R.~Romero, J.~Reynolds, E.~Mellers, A.~Redfern,
  L.~Lim, W.~Watts, C.~Patton, J.~Baker, {\em et~al.}, ``Design of monocular
  head-mounted displays for increased indoor firefighting safety and
  efficiency,'' in {\em Helmet-and head-mounted displays X: technologies and
  applications}, vol.~5800, pp.~103--114, SPIE, 2005.

\bibitem{lamon2019capability}
E.~Lamon, A.~De~Franco, L.~Peternel, and A.~Ajoudani, ``A capability-aware role
  allocation approach to industrial assembly tasks,'' {\em IEEE Robotics and
  Automation Letters}, vol.~4, no.~4, pp.~3378--3385, 2019.

\bibitem{dalle2021augmented}
M.~Dalle~Mura and G.~Dini, ``An augmented reality approach for supporting panel
  alignment in car body assembly,'' {\em Journal of Manufacturing Systems},
  vol.~59, pp.~251--260, 2021.

\bibitem{kosuge2004human}
K.~Kosuge and Y.~Hirata, ``Human-robot interaction,'' in {\em 2004 IEEE
  International Conference on Robotics and Biomimetics}, pp.~8--11, IEEE, 2004.

\bibitem{scholtz2003theory}
J.~Scholtz, ``Theory and evaluation of human robot interactions,'' in {\em 36th
  Annual Hawaii International Conference on System Sciences, 2003. Proceedings
  of the}, pp.~10--pp, IEEE, 2003.

\bibitem{lasota2017survey}
P.~A. Lasota, T.~Fong, J.~A. Shah, {\em et~al.}, ``A survey of methods for safe
  human-robot interaction,'' {\em Foundations and Trends{\textregistered} in
  Robotics}, vol.~5, no.~4, pp.~261--349, 2017.

\bibitem{AR}
J.~Carmigniani and B.~Furht, ``Augmented reality: an overview,'' {\em Handbook
  of augmented reality}, pp.~3--46, 2011.

\bibitem{mine1995virtual}
M.~R. Mine, ``Virtual environment interaction techniques,'' {\em UNC Chapel
  Hill CS Dept}, 1995.

\bibitem{cognolato2018head}
M.~Cognolato, M.~Atzori, and H.~M{\"u}ller, ``Head-mounted eye gaze tracking
  devices: An overview of modern devices and recent advances,'' {\em Journal of
  rehabilitation and assistive technologies engineering}, vol.~5,
  p.~2055668318773991, 2018.

\bibitem{holo}
Microsoft, ``{Microsoft} hololens 2.''
  https://www.microsoft.com/en-us/hololens/hardware, 2022.
\newblock Accessed: 2022-07-30.

\bibitem{joy}
T.~Kot, P.~Nov{\'a}k, and J.~Bajak, ``Using hololens to create a virtual
  operator station for mobile robots,'' in {\em 2018 19th International
  Carpathian Control Conference (ICCC)}, pp.~422--427, IEEE, 2018.

\bibitem{ARETT}
S.~Kapp, M.~Barz, S.~Mukhametov, D.~Sonntag, and J.~Kuhn, ``Arett: Augmented
  reality eye tracking toolkit for head mounted displays,'' {\em Sensors},
  vol.~21, no.~6, p.~2234, 2021.

\bibitem{Admittance}
C.~Ott and Y.~Nakamura, ``Admittance control using a base force/torque
  sensor.,'' {\em IFAC Proceedings Volumes}, vol.~42, no.~16, pp.~467--472,
  2009.

\bibitem{brooke1996sus}
J.~Brooke {\em et~al.}, ``Sus-a quick and dirty usability scale,'' {\em
  Usability evaluation in industry}, vol.~189, no.~194, pp.~4--7, 1996.

\bibitem{SEQ}
W.~Wetzlinger, A.~Auinger, and M.~D{\"o}rflinger, ``Comparing effectiveness,
  efficiency, ease of use, usability and user experience when using tablets and
  laptops,'' in {\em Design, User Experience, and Usability. Theories, Methods,
  and Tools for Designing the User Experience: Third International Conference,
  DUXU 2014, Held as Part of HCI International 2014, Heraklion, Crete, Greece,
  June 22-27, 2014, Proceedings, Part I 3}, pp.~402--412, Springer, 2014.

\end{thebibliography}
\end{document}